\documentclass[runningheads]{llncs}

\usepackage{times}
\usepackage{epsfig}
\usepackage{graphicx}
\usepackage{amsmath}
\usepackage{amssymb}
\usepackage{graphicx}
\usepackage{subfigure}
\usepackage{mathtools}
\usepackage{hyperref}
\usepackage{xcolor}
\UseRawInputEncoding

\usepackage{tikz,xcolor,hyperref}

\definecolor{lime}{HTML}{A6CE39}
\DeclareRobustCommand{\orcidicon}{%
	\begin{tikzpicture}
	\draw[lime, fill=lime] (0,0) 
	circle [radius=0.16] 
	node[white] {{\fontfamily{qag}\selectfont \tiny ID}};
	\draw[white, fill=white] (-0.0625,0.095) 
	circle [radius=0.007];
	\end{tikzpicture}
	\hspace{-2mm}
}

\foreach \x in {A, ..., Z}{%
	\expandafter\xdef\csname orcid\x\endcsname{\noexpand\href{https://orcid.org/\csname orcidauthor\x\endcsname}{\noexpand\orcidicon}}
}

\begin{document}
\title{Multiscale Detection of Cancerous Tissue in High Resolution Slide Scans}

% corresponding author: Corey Toler-Frankin corey.tolerfranklin@gmail.com 
% Other Author emails: 
% Qingchao Zhang  qingchaozhang@ufl.edu
% Coy D. Heldermon coy.heldermon@medicine.ufl.edu

\author{Qingchao Zhang \orcidA{} \and
Coy D. Heldermon\orcidB{} \and
Corey Toler-Franklin \orcidC{}}
\authorrunning{Q. Zhang et al.}
\institute{University of Florida, Gainesville FL 32611\\
\url{https://www.cise.ufl.edu}\\
\email{ctoler@cise.ufl.edu}}

\maketitle              % typeset the header of the contribution
%

%---------------------------------------------------
%-------------------------

\begin{abstract}
We present an algorithm for multi-scale tumor (chimeric cell) detection in high resolution slide scans. The broad range of tumor sizes in our dataset pose a challenge for current Convolutional Neural Networks (CNN) which often fail when image features are very small (8 pixels). Our approach modifies the effective receptive field at different layers in a CNN so that objects with a broad range of varying scales can be detected in a single forward pass. We define rules for computing adaptive prior anchor boxes which we show are solvable under the equal proportion interval principle. Two mechanisms in our CNN architecture alleviate the effects of non-discriminative features prevalent in our data - a foveal detection algorithm that incorporates a cascade residual-inception module and a deconvolution module with additional context information. When integrated into a Single Shot MultiBox Detector (SSD), these additions permit more accurate detection of small-scale objects.  The results permit efficient real-time analysis of medical images in pathology and related biomedical research fields.


\end{abstract}

%-------------------------

%-------------------------
\section{Introduction}

Deep learning algorithms have been effective for detecting and classifying metastatic cancer in medical images. Breast cancer detection in histopathology images~\cite{DBLP:journals/corr/BejnordiLGMGSKL17} and pulmonary lung cancer classification in CT scans~\cite{DBLP:journals/corr/abs-1712-05114} are examples which are essential for cancer diagnosis, staging and treatment~\cite{Kourou2015Machine}. CNN models are at the forefront of image-based classification methods that analyze high-resolution slide scans to detect tumors in tissue~\cite{Hou2016,yang2016,wang2016deep,Cruz2017Accurate}. These automated approaches are more efficient than manual methods or traditional supervised machine learning techniques that require hand-labeled annotations from practitioners with specialized expertise. 

We address challenges with CNN-based tumor (chimeric cell) detection and classification in medical datasets where tumor sizes vary significantly, and may be as small $8$ pixels. We propose a method to optimize the size and distribution of SSD~\cite{ssd} priors (anchor boxes) and adjust the receptive field to include more context. SSD is suited for tumor detection because it identifies multi-scale objects in one shot using information from multiple CNN layers. Priors, pre-computed bounding boxes that closely match the distribution of ground truth boxes, are used to define effective detection regions across scales. However, limitations with SSD make it less effective on microscopic scans in our domain. The range of prior anchor box sizes is fixed and limited. It is difficult to select the most effective prior anchor box parameters for a given dataset. Moreover,  results are inconsistent for less discriminating features (a known characteristic of pathology datasets).  Our approach increases the number of detectable scales by adaptively changing the aspect ratio of prior anchor boxes and modifying the receptive field to include a broader range of context and background information. Although additional context information has improved detection performance at deeper CNN layers in spatial recurrent neural networks~\cite{ION,fu2017dssd,kong2017ron}, the results have only been tested on nature scenes~\cite{imagenet_cvpr09}. We incorporate an iterative anchor box refinement algorithm~\cite{zhang2018single} that enhances performance.  The result is an effective small object detector~\cite{cao2018feature,DBLP:journals/corr/MengFCCT17} capable of locating tiny features. Compared to prior methods, our approach achieves higher detection rates for a broader range of object sizes in a single session. Our contributions include:

\begin{itemize}
\item Adaptive prior anchor boxes which we demonstrate are solvable under an equal proportion interval principle.
\item A foveal detection module that incorporates local context using cascade residual-inception modules. 
\item A deconvolution module that incorporates a broader range of background information.
\end{itemize}

%-------------------------

%-------------------------
\section{Related Work}

Object detection approaches related to our work can be divided into proposal-based and proposal-free frameworks. 

\paragraph{Proposal-based} methods are composed of proposal generation and classification stages. R-CNN~\cite{girshick2014rcnn} is an example that uses selective search~\cite{Uijlings13} and edge boxes~\cite{edge-boxes-locating-object-proposals-from-edges} for computing detection probabilities. Several modifications have been proposed  to improve speed. Fast R-CNN~\cite{girshick15fastrcnn} incorporates shared convolutional layers ~\cite{DBLP:journals/pami/HeZR015}. Faster R-CNN further increases accuracy and speed by replacing traditional proposal generation with proposal subnetworks~\cite{ren2015faster}. R-FCN~\cite{NIPS2016_6465} uses all convolutional layers and score maps for improved prediction results. Sematic segmentation-aware~\cite{DBLP:journals/corr/ZhuZLS16,gidaris2015object} and Mask R-CNN~\cite{he2017maskrcnn} approaches showed that context and segmentation integration improves detection accuracy. FPN~\cite{DBLP:journals/corr/LinDGHHB16} used pyramidal features to
detect multi-scale objects. HyperNet and Relation Networks utilize object relationships to improve accuracy~\cite{kong2016hypernet,hu2017}. 

\paragraph{Proposal-free} methods are faster than
two-stage proposal-based methods. The first successful proposal-free detector, YOLO~\cite{yolo,8100173}, applies a single neural network to the entire image. The image is automatically divided into regions for computing bounding box predictions. SSH~\cite{najibi2017ssh} uses integrated context information for face detection. SSD~\cite{ssd} is another example that utilizes multi-scale
information to boost both accuracy and speed in a single framework. SSD has been particularly successful for face detection~\cite{zhang3}. One key advantage of adopting SSD for tumor detection in medical images is it's ability to identify multiple size tumors in one session using multi-layer information. For this reason, SSD is extensively used to detect cancer in  CT~\cite{ssd_CT}, endoscopic~\cite{ssd_EI} and ultrasound images~\cite{ssd_UI}. Several SSD  variants integrate context information to increase accuracy~\cite{fu2017dssd,cao2018feature,ION,he2016deep}. We include pyramidal structures and adaptive prior anchors in a proposal-free approach. Although modifying SSD prior anchor boxes have been explored~\cite{kong2017ron,zhu2018seeing}, and refinements proposed for better detection performance~\cite{zhang2018single,zhang3,zhu2018seeing}, our approach is unique because we adaptively adjust prior anchor box aspect ratios, and provide anchor box distribution rules that  increase accuracy in microscopic slide scans. We also focus on optimizations for very small objects (which are often not detectable).

%-------------------------

%-------------------------
\section{Adaptive Prior Anchors} 
\label{sec:adaptiveAn}

\begin{figure}[h]
  \centering
  \includegraphics[width=0.8\linewidth]{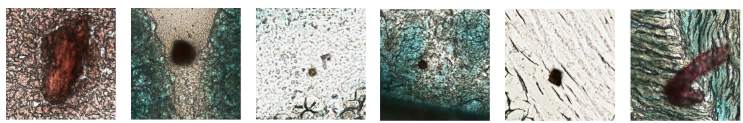}
  \caption{The size of image-based features in our patch samples range from $8-300$ pixels.  Chimeric maternal cell clusters in offspring.  Patch size: $300 \times 300$.}\label{fig:tumor-samples}
\end{figure}

We choose SSD as a starting point for our algorithm because, unlike scale-normalized detectors, it is better at detecting objects of multiple sizes. Figure~\ref{fig:tumor-samples} depicts the range of image-based feature sizes in our dataset of patches. In this section, we present details of our CNN architecture and SSD modifications.

\begin{figure}[h]
  \centering
  \includegraphics[width=0.6\linewidth]{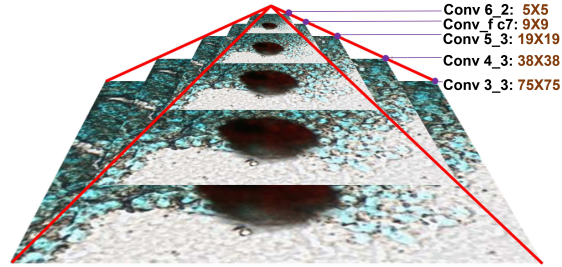}
  \caption{Our multiple feature maps and detection layers. }\label{fig:pyramid}
\end{figure}

\paragraph{High-resolution detection layers:} Our base convolutional network is VGG16 \cite{vgg}. The stride and receptive field size of layers in VGG16 increases with higher  layers \cite{yu2016visualizing}. However, the resolution decreases as shown in Figure~\ref{fig:pyramid}. The smaller objects in deeper layers will have much less information. For example, an object of size $32 \times 32$ pixels will only have an effective region of $2 \times 2$ pixles in $conv5\_3$. Therefore, we must rely on the shallow but high-resolution layers to detect small tumors. In our work, we add additional high-resolution $conv3\_3$ and $conv5\_3$ layers and remove SSD layers with strides that are too large. The details of our implementation are summarized in Table~\ref{tab:SSDtable}.

\begin{table}
       \centering
       \caption{Implementation parameters: detection layer, stride, anchor size, aspect ratio (AR) and receptive field (RF)%
       \label{tab:SSDtable}}

       \begin{tabular}{ccccc}
       \hline
       \textbf{Detection layer} & \textbf{Stride} & \textbf{Anchor size} & \textbf{Anchor AR} & \textbf{RF}\\
       \hline
       $conv3\_3$ & $4$ & $16$ & $1$, $2$, $\frac{1}{2}$ & $48$\\
       $conv4\_3$ & 8 & 32 & 1, $\frac{3}{2}$, 3, $\frac{2}{3}$, $\frac{1}{3}$ & 108\\
       $conv5\_3$ & 16 & 64 & 1, $\frac{3}{2}$, 3, $\frac{2}{3}$, $\frac{1}{3}$ & 228\\
       $conv\_fc\_7$ & 32 & 128 & 1, $\frac{3}{2}$, 3, $\frac{2}{3}$, $\frac{1}{3}$ & 340\\
       $ conv6\_2$ & 64 & 256& 1, $\frac{3}{2}$, $\frac{2}{3}$& 468\\
       \hline
       \end{tabular}
       \end{table}

\begin{figure}[h]
       \centering
       \subfigure[SSD]{
       \label{fig:distributiona} %% label for first subfigure
       \includegraphics[width=0.4\linewidth]{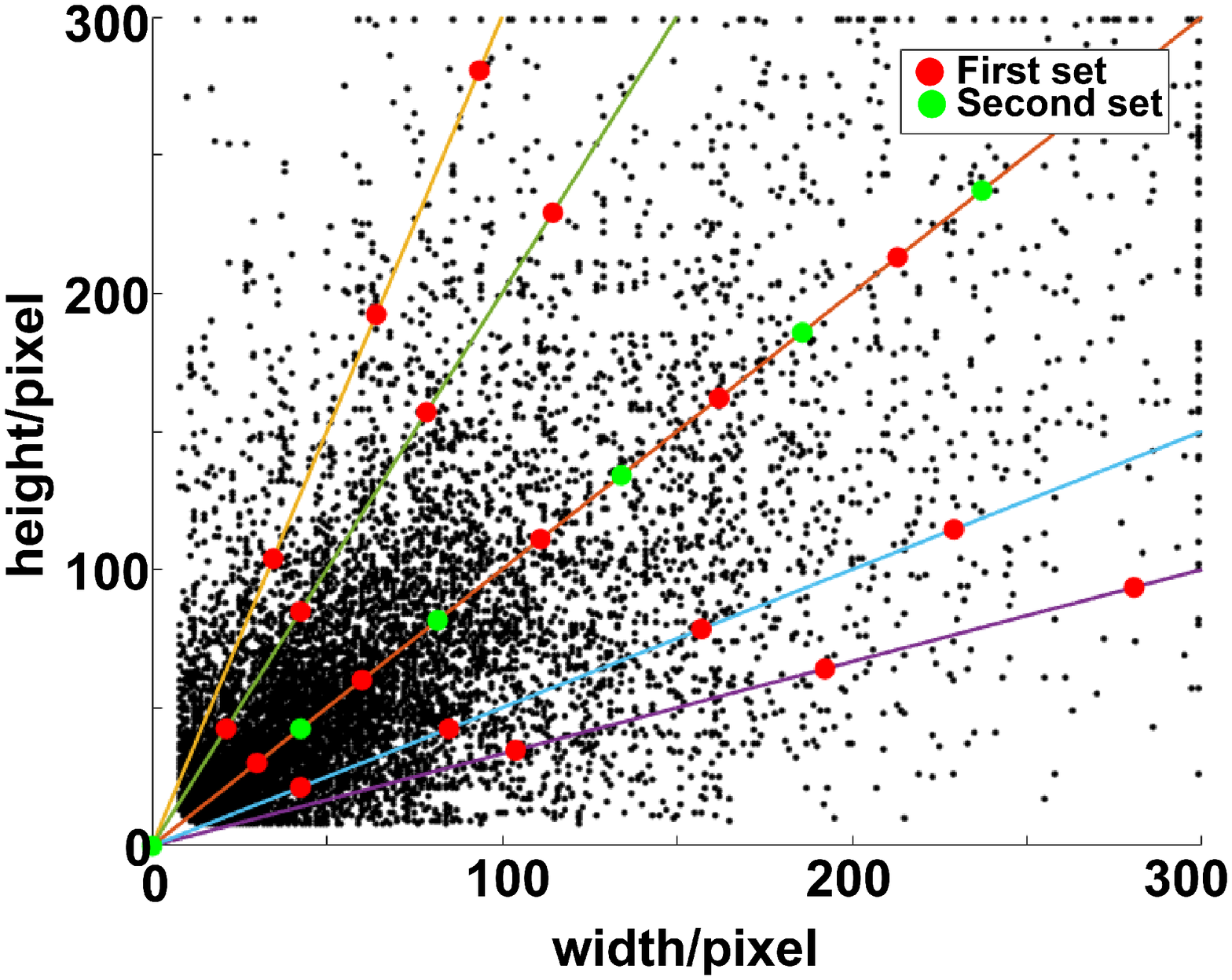}}
       %\hspace{1in}
       \subfigure[Our Approach]{
       \label{distributionb} %% label for second subfigure
       \includegraphics[width=0.4\linewidth]{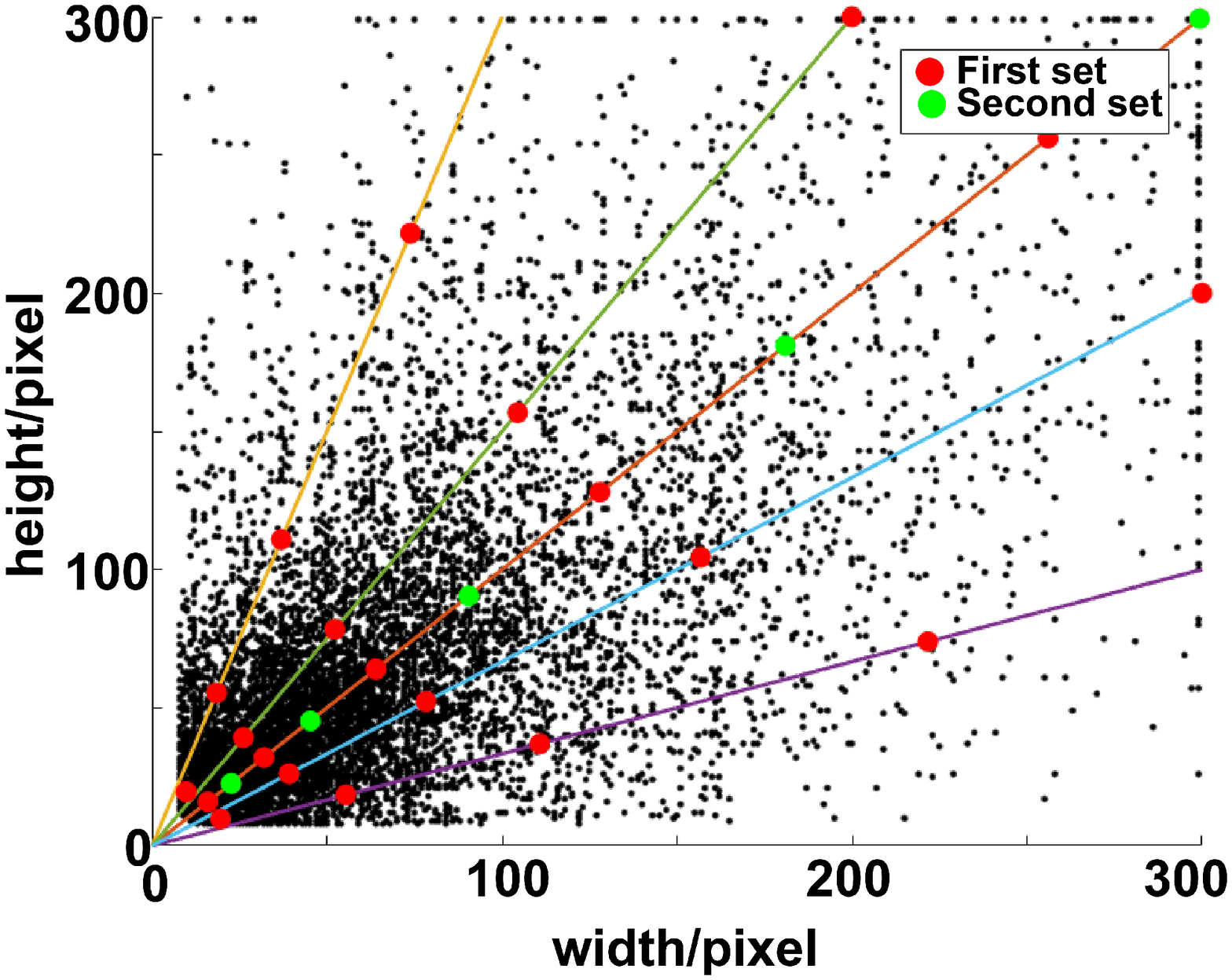}}
       \caption{The point cloud (black) represents the distribution of the tumor width and height. The colored dots denote the width and height of detection anchors. The colored lines connect anchors with the same AR but different scale. }
      \label{fig:distribution} %% label for entire figure
      \end{figure}

\paragraph{Equal-proportion interval anchors:} The effective receptive field is smaller than the theoretical receptive field~\cite{luo2016understanding}. Therefore, anchors in each layer should be smaller than the responding theoretical receptive field. We adopt the equal-proportion interval principle~\cite{zhang3} which insures that anchors possess equal density compared to other methods (like the regular-space method in SSD). Comparing Figures~\ref{fig:distributiona} and~\ref{distributionb}, equal-proportion interval anchors are better suited for multi-scale detection given the  distribution of tumors in our images (more anchors at smaller but denser tumor regions). In addition, anchors that conform to the equal-proportion rule are more easily integrated into the receptive field and stride. This is because the  receptive field and stride in SSD detection layers increase at a near proportional rate. Table~\ref{tab:SSDtable} shows our anchor design.

\paragraph {Aspect Ratio Design:} The aspect ratio (AR) defines the anchor profile. Here we adopt the width and height relationship:

\begin{equation}
W_{anchor} = size\cdot\sqrt{AR}
\end{equation}

\begin{equation}
H_{anchor} = \frac{size}{\sqrt{AR}}
\end{equation}

\noindent All previous anchor-based object detection frameworks~\cite{Ren:2015:FRT:2969239.2969250,ssd,zhang3} use a subjective \emph{rule of thumb} to determine AR. \textbf{We propose an AR design criterion}. (1) AR values should be as small as possible to reduce the total number of anchors. (2) AR values must be large enough to cover almost every object (at least $99\%$) to have a good recall. (3) AR values may differ from layer-to-layer depending on the detection criteria of the layer. In Section~\ref{sec:AR}, we give a mathematical proof that the maximum AR (mAR) of anchors is only relevant to the threshold of intersection-over-union(IoU) and mAR of objects if the anchor of different layers are designed by the equal-proportion interval principle. The maximum anchors' AR ($mAR_{anchor}$) in each layer shall be chosen by Equation~\ref{equ:AR}
       
       $ mAR_{anchor} = $
       \begin{equation}\label{equ:AR}
       \begin{split}
         mAR_{obj} \cdot max\{(\frac{2}{1+\frac{1}{IoU}})^2, \ \frac{IoU}{2-2IoU},\ IoU\}
         \end{split}
       \end{equation}
       where $mAR_{obj} = sup_{i \in \Omega} \{AR_{obj}^i\} $, $AR_{obj}$ is the objects' AR.
       %The $mAR_{anchor}$ shall increase with the increase of $mAR_{obj}$, and decrease with the increase of $IoU$.\\
       
According to our statistics, $99\%$ of the AR of the samples in our tumor datasets are less than $6$ pixels. Most of the exceeding $1\%$ of samples result from bad cropping (i.e. only cutting edges of the stained region). Therefore, we select the $mAR_{obj}$ as $6$. We choose $IoU$ as $0.5$ which is standard in prior work~\cite{girshick2014rcnn,ssd,Ren:2015:FRT:2969239.2969250}. According to Equation~\ref{equ:AR}, we select $mAR_{anchor} = 3$. Having more anchors often weakens performance~\cite{thesis}. Thus, we just choose a discrete $AR_{anchor} \in \{ 1, 1.5, 3\}$ to control the number of anchors. Compared to SSD ($AR_{anchor} \in \{ 1, 2, 3\}$)~\cite{ssd} depicted in Figure~\ref{fig:distributiona}, our AR design covers more objects in Figure~\ref{distributionb} and has a better recall.
We also observed  that smaller tumor regions tend to be round (with smaller $AR_{obj}$). Thus the $AR_{anchor}$ of the first layer can be chosen to be smaller, to decrease the total number of anchors and reduce computation cost. The $AR_{anchor}$ of the last layer must not exceed the  size of the whole slide image. Finally, the $AR_{anchor}$ of the $conv3\_3$ and $conv6\_2$ are chosen as $\{1, 2\}$ and $\{1, 1.5\}$ respectively. The above calculation only considers  the $width \geq height$ case due to symmetry. The final design params are in Table~\ref{tab:SSDtable}.

\paragraph{Double sets of anchors:} Although our anchors cover almost all ground truth (GT) boxes in the above analysis, different box sizes still correspond to different numbers of anchors. Statistics~\cite{zhang3} show tiny outer objects are less likely to have a suitable anchor match. In order to alleviate this problem, we adopt double sets of anchors. We denote $\{S_1, ... S_k, ...\}$ as the first set anchors and $\{S'_1, ... S'_k, ...\}$ as the second set anchors. The size and AR values of the first set of anchors are shown in Table~\ref{tab:SSDtable}. The second set of anchors are $ S'_k = \sqrt{S_k\cdot S_{k+1}}$ with aspect ratio $1$, except for $ S'_{conv6\_2}$ which is equal  to the image-patch size $300$~\cite{ssd}. The final anchor sizes are depicted in Figure~\ref{distributionb}.

%-------------------------

%-------------------------
\section{Adaptive Aspect Ratio} \label{sec:AR}

\begin{figure}[h]
  \centering
  \includegraphics[width=0.5\linewidth]{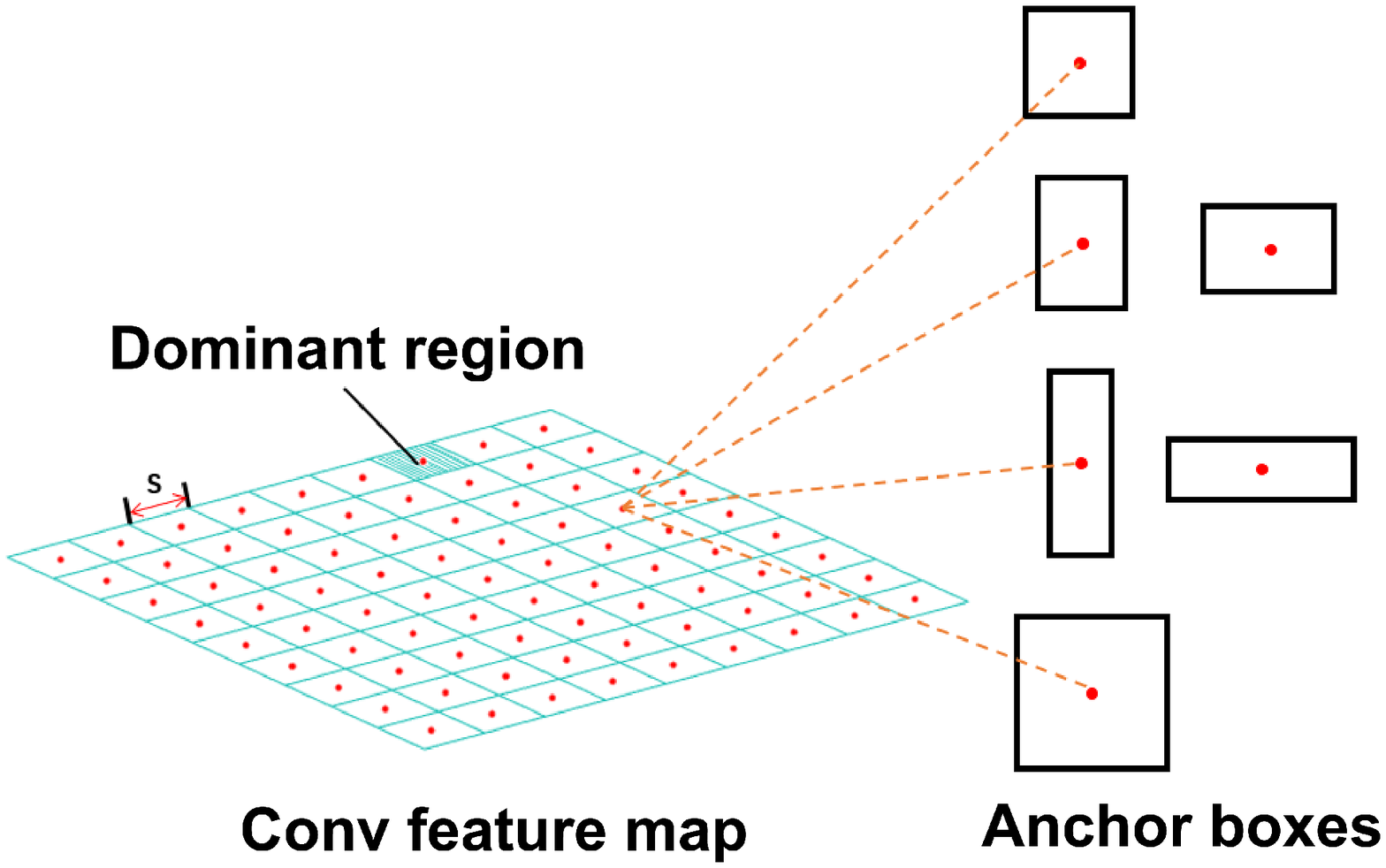}
  \includegraphics[width=0.3\linewidth]{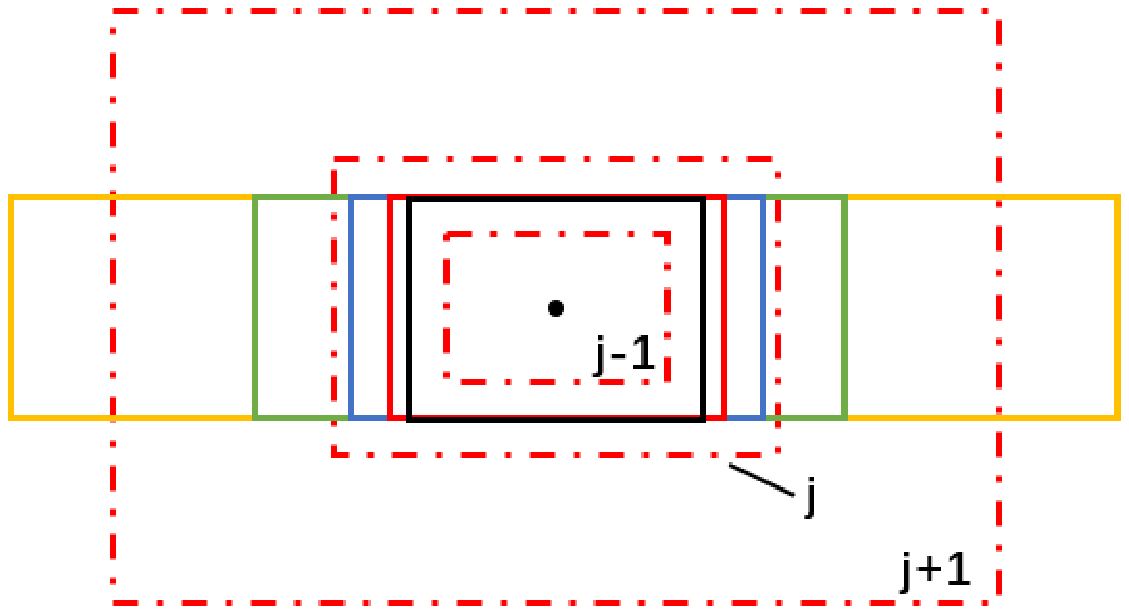}
  \caption{\textbf{Left:} (left) Feature map ($conv\_fc\_7$). Boxes in the grid (cyan) are dominant regions with side length equal to stride $S$. The dotted red lines connect to anchor boxes on this feature layer. (right) The layout of homocentric GT boxes and anchors. Dashed boxes are equal-proportion anchors with equal AR. Solid boxes are GT boxes where the red one has the same AR as the anchors.}\label{fig:dominant}
\end{figure}

\paragraph{Anchor Boxes:} Figure~\ref{fig:dominant} illustrates an example of anchor boxes in a layer $j$. Anchor boxes are rotation-invariant. The center of each box is aligned at a distance $\delta$, where $\delta = S_{j}$, the stride of $j^{th}$ layer. We define a dominant area $C_x \pm \frac{1}{2}S_j, C_y \pm \frac{1}{2}S_j$ where $(C_x, C_y)$ is the center of each anchor. The highest $IoU$ for an object occurs where the anchor is centered   closest to it's ground truth center~\cite{DBLP:journals/corr/abs-1802-09058}. This means that the $IoU$ will reach a maximum for objects with GT boxes in the dominant region of the anchor while objects with centers outside the dominant region will be matched with a neighboring anchor. As shown in Table~\ref{tab:SSDtable}, $S_j$ is much smaller than the anchor box size. For simplicity, we presume the object and anchor are concentric if matched.

The size of the anchor is chosen by the equal-proportion interval principle $\{$ $2^f$, $2^{f+1}$, ..., $2^{j}$, ...,$2^l$$\}$, where $f$ is the first detection layer and $l$ the last. For convenience, we only consider symmetrical cases where $width \geq height$. Here, we denote $t$ as the maximum AR of anchors in the $j^{th}$ layer.
 We presume the width ($w$) and height ($h$) of objects follows ( $w \leq kh$ ), where $k$ is the maximum object AR, i.e.\ $k = sup \{\frac{w_i}{h_i}\}$. Note that $IoU$ satisfies $IoU \geq T$, where $T$ is a constant threshold. In our analysis, we only consider objects where AR ( $w = kh$ ). By satisfying the maximum AR, we satisfy all remaining objects and conditions.
 
 We set $h$ to lie in the interval between $(j-1)^{th}$ and $j^{th}$ set anchor (i.e.\ $\frac{2^{j-1}}{\sqrt{t}} < h \leq \frac{2^j}{ \sqrt{t}}$), as shown in Figure~\ref{fig:dominant} \emph{right}. There are only four cases of $w$: \textbf{1)} $w > 2^{j+1}\sqrt{t}$; \textbf{2)} $2^j\sqrt{t} < w \leq 2^{j+1} \sqrt{t}$; \textbf{3)} $ht < w \leq 2^j \sqrt{t}$; \textbf{4)} $w \leq ht$, corresponding to yellow, green, blue and black solid boxes in Figure~\ref{fig:dominant} \emph{right} respectively. The first three cases correspond to case \textbf{(1) $k\geq t$}, the last corresponds to \textbf{(2) $k< t$}. Below we will discuss how to obtain $t$ per these two cases.

\begin{figure}[h]
  \centering
  \includegraphics[width=0.6\linewidth]{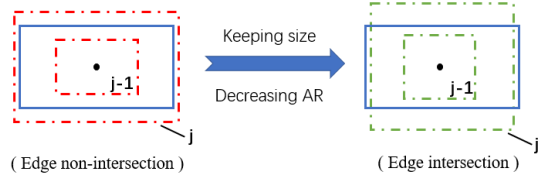}
  \caption{The edge non-intersection case ($AR = t$) between $j^{th}$  anchor and GT box can be transformed to edge-intersection ($AR = t'$) by decreasing anchors' AR while keeping size consistent. The dashed boxes are anchors, the solid boxes are object GT boxes. }\label{jjj2}
\end{figure}

{Case \textbf{(1)}}, $k\geq t$. \\
 There are three cases in first case: \textbf{1)} $w > 2^{j+1}\sqrt{t}$; \textbf{2)} $2^j\sqrt{t} < w \leq 2^{j+1} \sqrt{t}$; \textbf{3)} $ht < w \leq 2^j \sqrt{t}$.
 \\ Case \textbf{1) $\&$ 2)}: $w > 2^j\sqrt{t}$
 \\The IoU between Gt and the $(j-1)^{th}$ anchor reaches maximum only if $h \rightarrow \frac{2^{j-1}}{\sqrt{t}}$ $\&$ $w \rightarrow 2^j \sqrt{t}$.
 Then
 \begin{equation}\label{la}
   max(IoU_{j-1}) \rightarrow \frac{\frac{2^{j-1}}{\sqrt{t}} \cdot 2^{j-1} \sqrt{t}} { \frac{2^{j-1}}{\sqrt{t}} \cdot 2^j \sqrt{t}} = \frac{1}{2}
 \end{equation}
 Simultaneously,
  \begin{equation}\label{la2}
   IoU_{j} \rightarrow \frac{\frac{2^{j-1}}{\sqrt{t}} \cdot 2^{j} \sqrt{t}} { \frac{2^{j}}{\sqrt{t}} \cdot 2^j \sqrt{t}} = \frac{1}{2}
 \end{equation}
 Therefore, we conclude $IoU_{j-1} \leq IoU_{j}$ under Case \textbf{1) $\&$ 2)}. Similarly, we know $IoU_{j+1} \leq IoU_{j}$. Thus, $IoU_{j}$ is the largest one under this condition.
 \\ Case \textbf{3)}: $ht < w \leq 2^j \sqrt{t}$
 \\In this case, the GT box is totally enclosed by the $j^{th}$ anchor, as the edge non-intersection in Fig. \ref{jjj2}. We can transform Case \textbf{3)} to Case \textbf{1) $\&$ 2)} by replacing $t$ by a smaller $t'$ while keeping size, as illustrated in Fig. \ref{jjj2}. Note that $t'$ is a better solution as $t' < t$.

 Given the above analysis, we conclude that $IoU_j$ is the largest under \textbf{Case (1)}. Consequently, we assign the ground truth box ($w \times h$) to anchors in the $j^{th}$ layer if $\frac{2^{j-1}}{\sqrt{t}} < h \leq \frac{2^j}{ \sqrt{t}}$. Otherwise, if $h \leq \frac{2^{j-1}}{\sqrt{t}}$, it will be assigned to the $(j-1)^{th}$ layer, else if $h > \frac{2^j}{ \sqrt{t}}$, it will be assigned to the $(j+1)^{th}$ layer.
\\According to the above analysis, we can formulate the following equations:
 \begin{eqnarray}\label{aaa}
   IoU &=& \frac{I}{U} \quad \geq \quad T \\
   I &=& h \cdot 2^j \sqrt{t} \\
   U &=& (w - 2^j \sqrt{t}) h + 2^j \sqrt{t} \cdot \frac{2^j}{\sqrt{t}} \\
   w &=& kh
 \end{eqnarray}
 where the $I$ denotes the intersection and $U$ denotes the union.\\
 Solving the above array produces Equation~\eqref{dd}
 \begin{equation}\label{dd}
   Tkh^2 - (T+1)2^j\sqrt{t}h + T2^{2j} \leq 0
 \end{equation}
 If Equation~\eqref{dd} has a solution, it must satisfy $\Delta \geq 0$, where $\Delta$ is
 \begin{equation}\label{cc}
   \Delta  = 2^{2j} [(T+1)^2 t - 4T^2k]
 \end{equation}
From Equation~\eqref{cc}, we can generate the range of $t$:
\begin{equation}\label{t}
  t \geq (\frac{2}{1+\frac{1}{T}})^2\cdot k
\end{equation}
Equation~\eqref{dd} can be written in form:
\begin{equation}\label{fh}
  f(h) \leq 0, \qquad f(h) = Tkh^2 - (T+1)2^j\sqrt{t}h + T2^{2j}
\end{equation}
Because $f(h)$ is a convex function and $h \in (\frac{2^{j-1}}{\sqrt{t}}, \frac{2^j}{ \sqrt{t}}]$, the necessary condition of Equation~\eqref{fh} is $f(\frac{2^{j-1}}{\sqrt{t}}) \leq 0$ and $f(\frac{2^{j}}{\sqrt{t}}) \leq 0$:
\begin{equation}\label{fh-left}
  f(\frac{2^{j-1}}{\sqrt{t}}) = Tk\frac{2^{2j-2}}{t} - (T+1) 2^j \sqrt{t} \cdot \frac{2^{j-1}}{\sqrt{t}} + T2^{2j} \leq 0
\end{equation}
\begin{equation}\label{fh-right}
  f(\frac{2^j}{\sqrt{t}}) = Tk \frac{2^{2j}}{t} - (T+1) 2^j \sqrt{t} \cdot \frac{2^j}{\sqrt{t}} + T2^{2j} \leq 0
\end{equation}
Solving Equations~\eqref{fh-left}~\eqref{fh-right} respectively results in:
\begin{equation}\label{eq:1}
  t \geq \frac{Tk}{2-2T}
\end{equation}
and
\begin{equation}\label{eq:2}
  t \geq Tk
\end{equation}
We would like $t$ to be as small as possible to reduce the total number of anchors, and from Equations~\eqref{t},~\eqref{eq:1} and~\eqref{eq:2}, we formulate:
\begin{equation}\label{eq:3}
  t = max\{(\frac{2}{1+\frac{1}{T}})^2k,\ \frac{Tk}{2-2T},\ Tk\}
\end{equation}
We can see that because $T \in [0, 1]$, so $t \in (0, k]$. The condition $k\geq t$ always holds.
\\From our statistics, we can find that $k = 6$. If $T$ is simply chosen to be $0.5$, we can find that $t = 3$ from Equation~\eqref{eq:3}.
\paragraph{Case \textbf{(2)}}$k< t$.
\\From the analysis in \textbf{(1)}, we draw that for any $T \in [0, 1]$, there exist a $t'$ satisfy Equation~\eqref{t} and $t' \leq k < t$. Just replacing $t$ with $t'$, we can produce a better result, as illustrated in Fig. \ref{jjj2}. Here, for our specific problem (i.e. $k = 6$, $T = 0.5$), we use another simpler method to prove it.
\\From $\frac{2^{j-1}}{\sqrt{t}} < h \leq \frac{2^j}{ \sqrt{t}}$ and $k< t$, results in: \begin{equation}\label{qq}
  w = kh < th  \leq t \cdot \frac{2^j}{ \sqrt{t}} = 2^j \sqrt{t}
\end{equation}
where $2^j \sqrt{t}$ is the anchor width. Thus, both $w$ and $h$ are less than the width and height in $j^{th}$ layer, which we denote as $W_j$ and $H_j$ respectively. Similarly, we have $w > W_{j-1}$ and $h>H_{j-1}$. So, we have
\begin{equation}\label{z}
  W_{j-1} < w < W_j
\end{equation}
\begin{equation}\label{z2}
  H_{j-1} < h < H_j
\end{equation}
where, from the anchor equal-proportion interval principle, $W_j = 2W_{j-1}$ and $H_j = 2H_{j-1}$. Equation~\eqref{z} $\times$ Equation~\eqref{z2}, we can generate:
\begin{equation}\label{z3}
  A_{j-1} < a < A_{j}
\end{equation}
where, $A_{j-1}$ and $A_j$ are areas of anchor $j-1$ and $j$, and $a = h \cdot w$. This results in $A_{j} = 4A_{j-1}$.
Now, if $A_{j-1} < a \leq 2A_{j-1}$, the $IoU$ between the GT and $(j-1)^{th}$ layer anchor will be $\frac{A_{j-1}}{a} \geq 0.5$. Similarly, we find $\frac{a}{A_{j}} > 0.5$ if $2A_{j-1} < a < A_{j}$. Therefore, the ground truth box will have a $IoU$ not less than $0.5$ with anchors either in $(j-1)^{th}$ or $j^{th}$ layer. Thus, we claim that the GT boxes will always have an anchor to match, if the condition $k< t$ holds. Actually, this claim holds for any Edge non-intersection cases (i.e. $w \leq 2^{j} \sqrt{t}$) with $j^{th}$ layer in Figure~\ref{fig:dominant} \emph{right}.

%-------------------------

%-------------------------
\section{Foveal Context-Integrated Detection}\label{sec:foveal}

Tumor detection in medical images is often more challenging than applications like face detection which rely on color and texture patterns that are more discriminative. As shown in Figure~\ref{fig:tumor-samples}, there is a broad variation in feature types (scale, shape, color). Moreover, tumor (chimeric cell) detection depends on a number of factors which are not limited to image-based features like color. Background context is important, particularly for dense features with little to no color variations. To address these challenges, we now introduce a context integrated detection module, similar to a foveal structure~\cite{zagoruyko2016multipath}.

%\input{../fig-origin}
 % \begin{figure}[t]
%  \centering
  %\usepackage{graphicx}
%  \includegraphics[width=0.1\linewidth]{./images/tumor.eps}
%  \caption{The multiple receptive field of our foveal residual-deconv-%inception prediction module. }\label{fig:tumor}
%\end{figure}

\begin{figure}[h]
       \subfigure[]{
       \label{fig:origin} %% label for first subfigure
       \includegraphics[width=0.2\linewidth]{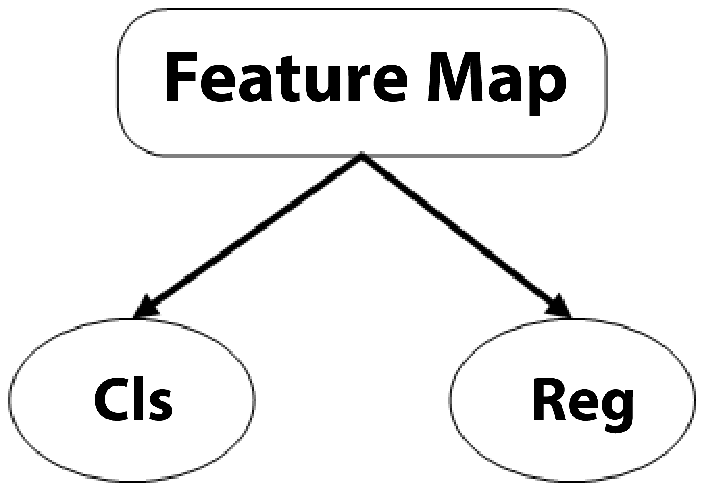}}
       %\hspace{1in}
       \subfigure[]{
       \label{fig:origin-inception} %% label for second subfigure
       \includegraphics[width=0.2\linewidth]{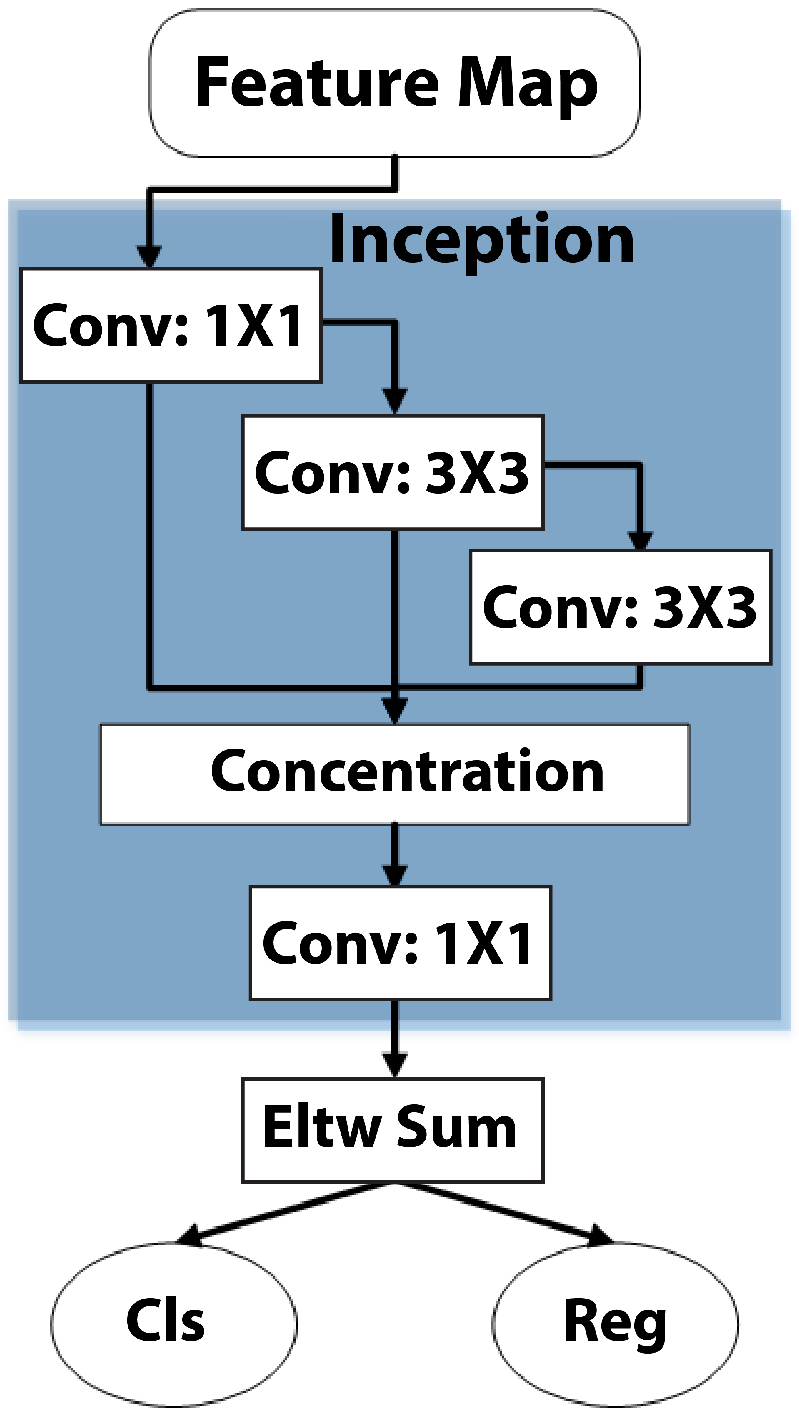}}
       %\hspace{1in}
       \subfigure[]{
       \label{fig:inception} %% label for second subfigure
       \includegraphics[width=0.2\linewidth]{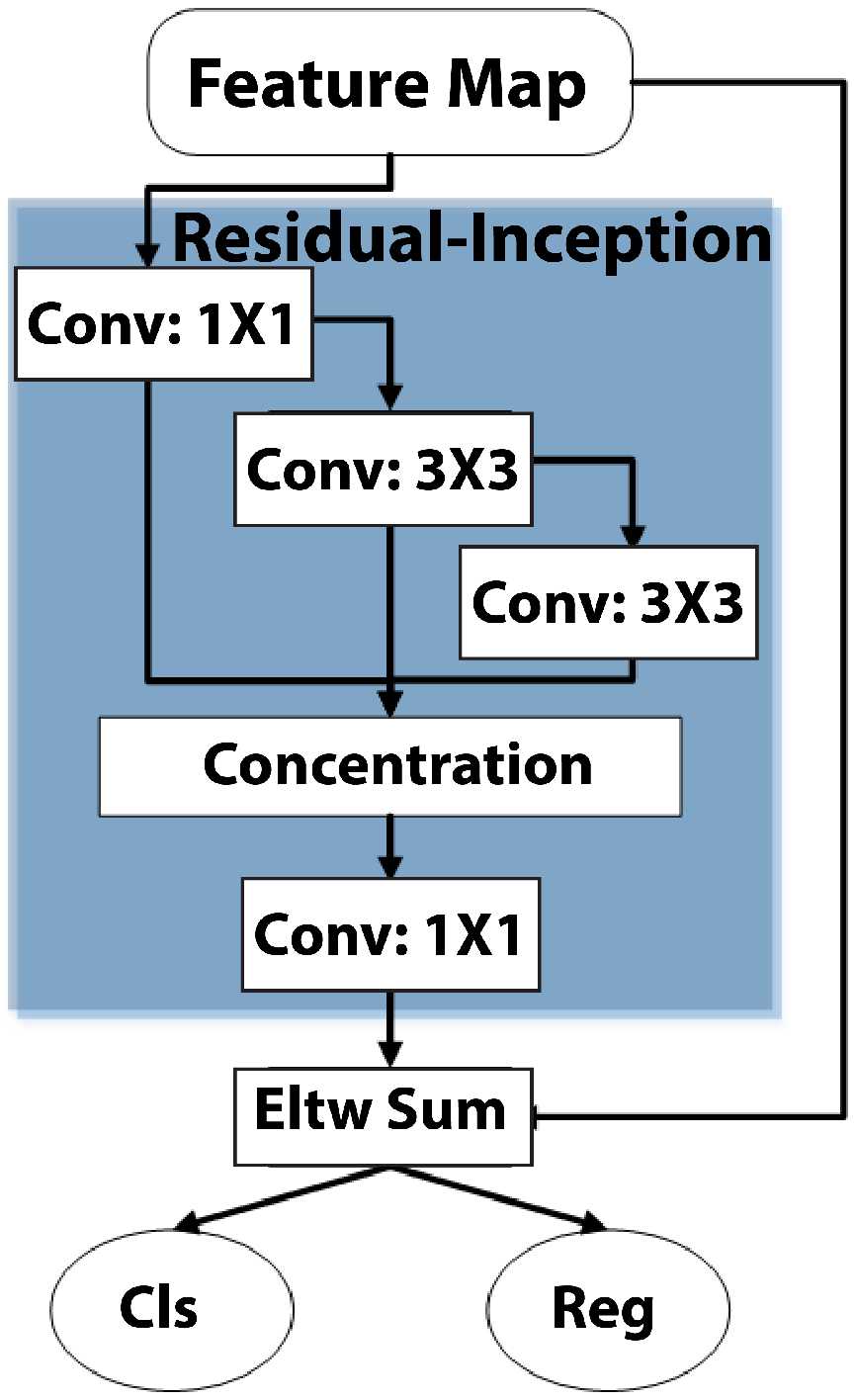}}
       %\hspace{0.5in}
       \subfigure[]{
       \label{fig:inception-deconv} %% label for second subfigure
       \includegraphics[width=0.3\linewidth]{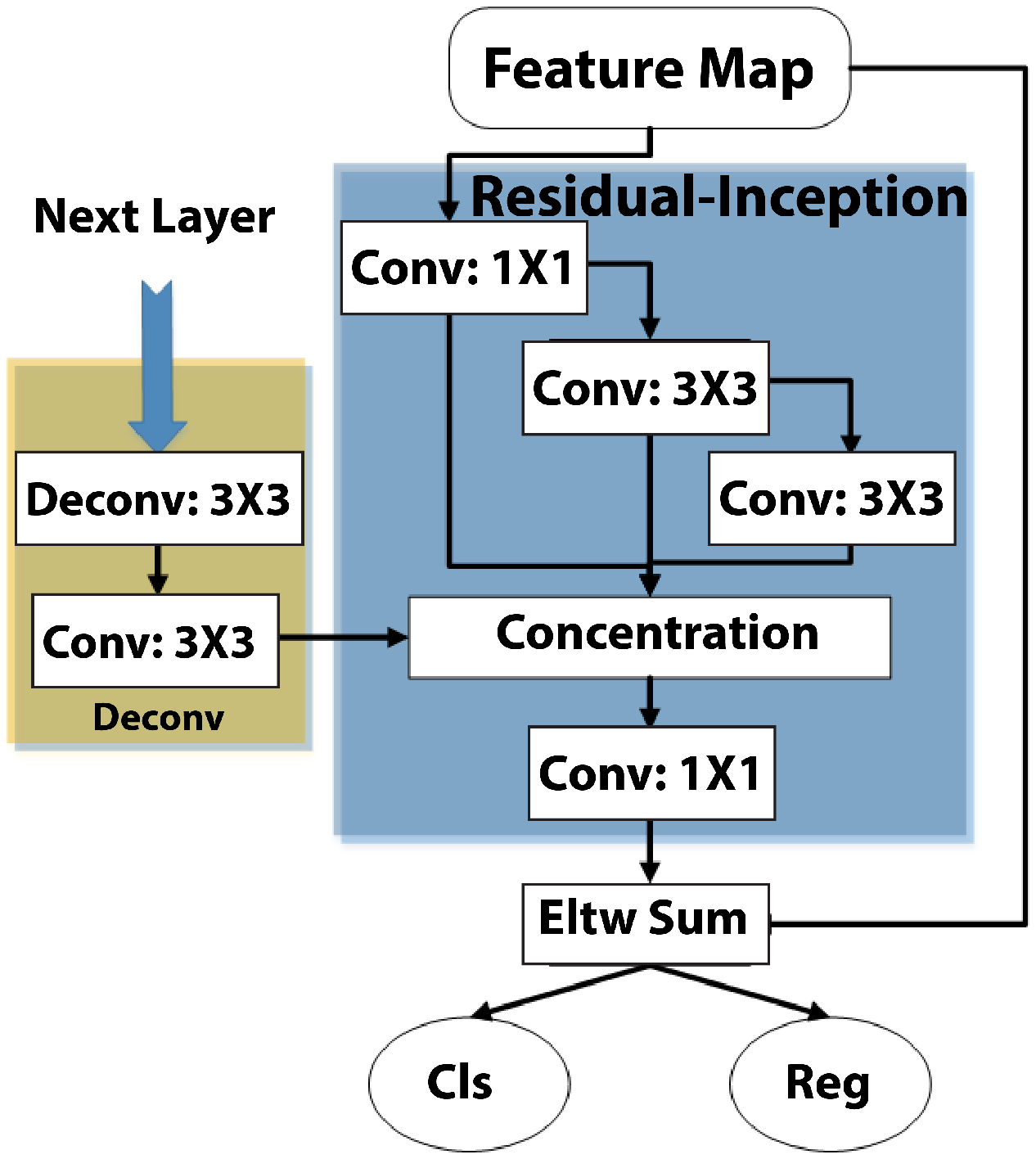}}
       \caption{Multiple prediction modules. Cls: classification. Reg: localization regression.~\ref{fig:origin} SSD, SSD adaptive:~\ref{fig:origin-inception}inception,~\ref{fig:inception} residual-inception~\ref{fig:inception-deconv}, residual-deconvolution.}
      \label{fig:inception22} %% label for entire figure
      \end{figure}

\paragraph{Cascade residual-inception module:} The Inception block~\cite{DBLP:journals/corr/KimCHRP16} provides multiple receptive fields per chosen convolutional kernel size~\cite{szegedy2015going}. Inspired by work in this area~\cite{fu2017dssd,lee2017wide}, we make a separate residual-inception prediction module instead of predicting directly on the feature layer (Figure~\ref{fig:origin}). Residual-inception is an effective combination of the inception module and the residual block~\cite{he2016deep}. We achieve improvement over prior examples~\cite{lee2017wide,szegedy2015going} by altering the parallel convolutional layers of the inception module to form a cascade structure. Each convolutional layer in the cascade shares computation instructions and memory. Figure~\ref{fig:inception} outlines our implementation. We have $1 \times 1, 3 \times 3$ and $5 \times 5$ ( by $2$ cascade $3 \times 3)$ conv kernels. The first $1 \times 1$ kernel reduces the dimension of feature map (i.e. the channel number). The cascade residual-inception module generates multiple receptive fields at lower computational cost. Our performance is similar to current inception modules~\cite{szegedy2015going} with only $25\%$ of the parameters. Experimental results show that this design converges faster and produces a higher accuracy rate.

\paragraph{Deconvolution module with context information:}
To utilize broader context information, we adopt the deconvolution shown in Figure~\ref{fig:inception-deconv}. Other approaches like DSSD~\cite{fu2017dssd} incorporate context elements naively by adding or multiplying elements in the deconvolution feature map with the original feature map. This does not work well with our residual-inception structure because it reduces performance. We adopt the feature-fusion concatenation method~\cite{cao2018feature}. This permits the convolutional layer to effectively learn useful background information while decreasing interference from noise. We concatenate the deconvolution module into the inception module as shown in Figure~\ref{fig:inception-deconv}. This design integrates hierarchical context information. The inception module provides local context while the deconvolution module provides broader context information.

In our implementation, we only add the foveal context-integrated module on the first three SSD detection layers where there is a small receptive field and higher resolutions. These shallow layers are mainly responsible for detecting small objects~\cite{zhang3}. Large objects rely  much less on context compared to small objects~\cite{hu2017finding}. Thus, we only add the foveal detection module to shallow layers so that  small objects are more recognizable.

%-------------------------

%-------------------------
\section{Experimental Evaluation}

We evaluate our system on a dataset of $30,000$  patches extracted from microscopic scans (Figure~\ref{fig:tumor-samples}). Each patch is $300 \times 300$ pixels. Tumor sizes in our scans range from $8-300$ pixels. We separate patches into training and testing sets ($9 : 1$ respectively). We generate precision recall and average precision values by comparing the detected tumor regions with hand-labeled ground truth images. Our code is implemented using the \emph{Caffe} toolbox~\cite{jia2014caffe} on  a \emph{Windows 10} system. We trained using a NVIDIA GTX 1080Ti GPU for $3$ days with $2$ additional days for finetuning. Training and finetuning was performed for $180,000$ and $100,000$ iterations respectively. 

\begin{figure}[h]
\centering
  \includegraphics[width=0.97\linewidth]{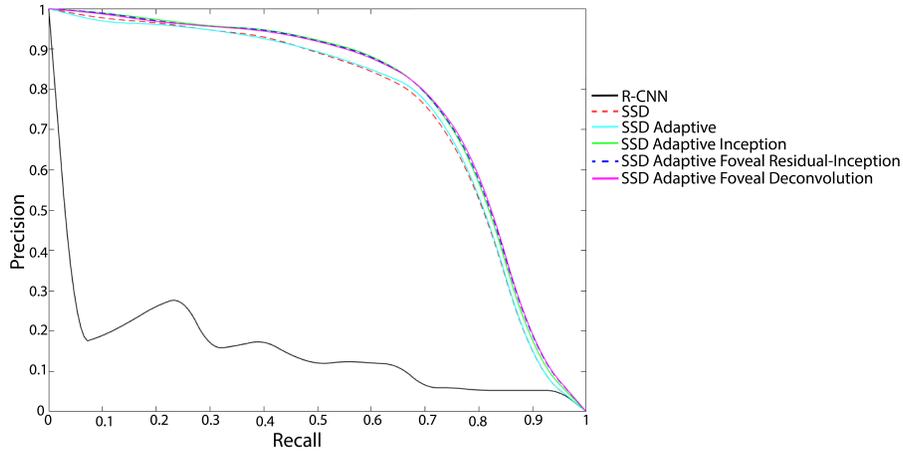}
  \caption{Our SSD modifications produce higher precision for a longer recall when detecting multi-scale tumors in high resolution slide scans.}\label{fig:pr}
\end{figure}

Table~\ref{tab:result} compares our approach with R-CNN and traditional SSD. Our modifications - SSD adaptive, SSD adaptive foveal residual-inception and SSD  adaptive foveal deconvolution – have a higher recall (0.699,  0.699, 0.702 respectively) compared to R-CNN and traditional SSD  (0.713 and 0.608  respectively) as we detect small tumors that are undetectable in other methods. Figures~\ref{fig:distributiona} and~\ref{distributionb} show tumor and anchor box size distributions for SSD and our method. Our adaptive prior anchors produce higher average precision rates~\cite{Everingham15}  (0.747, 0.7485 and 0.7497 respectively) compared to R-CNN and traditional SSD (0.175 and 0.726 respectively). The precision-recall curves in Figure~\ref{fig:pr} show that SSD-based approaches have a higher precision for a longer recall than R-CNN. Our modifications outperform traditional SSD. Although proposal-based R-CNN has a high recall, it's precision is low as most candidate boxes are ineffective in our domain. The two-stage process is also inefficient. SSD has been shown to out-perform faster R-CNN while Mask R-CNN requires segmentation  and is not popular for pure bounding box object detection. Our reliance on patch extraction to resolve GPU memory restrictions is a limitation which increases total computation time. Although SSD is more efficient with higher precision, there is no built-in guarantee anchor distributions will be optimal, and no mechanism to tackle non-discriminative tumors.

\begin{table}[h]
       \centering

       \caption{Comparing R-CNN, SSD and our modified SSD (adaptive, adaptive foveal residual-inception and adaptive foveal deconvolution). Our design yields the best results per average precision (AP). (precision recall threshold = $0.5$)%
       \label{tab:result}}

       \begin{tabular}{cccc}
       \hline
       \textbf{Method} & \textbf{Precision}& \textbf{Recall} & \textbf{AP}\\
       \hline
       R-CNN & 0.065 & 0.713 & 0.175 \\
       SSD & 0.826 & 0.644& 0.726 \\
       SSD-adaptive & 0.808 & 0.699 & 0.747\\
       SSD-adaptive foveal residual-inception & 0.809 & 0.699 & 0.7485\\
       SSD-adaptive foveal deconvolution & 0.809 & 0.702 & 0.7497\\
       \hline
       \end{tabular}
       \end{table}

%-------------------------

%-------------------------
\section{Conclusion}

We presented a multiscale detection algorithm that augments SSD for successful detection of small tumors (chimeric cells) in patches extracted from high-resolution slide scans. Our approach includes a modified SSD deconvolution module that integrates background information and a foveal context module with a cascade residual inception module for local context integration. We introduced rules for computing adaptive prior anchor boxes with aspect ratios and distributions that are more suitable for detecting small-scale objects ($8$ pixels). Our evaluation methods show that our approach produces a higher precision recall when  compared with traditional SSD. Our method is also effective for non-discriminative feature sets.

%-------------------------

%-------------------------
\vspace{10pt}
\noindent\textbf{Acknowledgements.} We thank K08 DK085141 and the American Cancer Society Chris DiMarco Institutional Research Grant (CDH) for funding support. This work was conducted at the UF Graphics Imaging and Light Measurement Lab (GILMLab).

%-------------------------

%---------------------------------------------------

\bibliographystyle{splncs04}
\bibliography{egbib}

\end{document}